\documentclass{IEEEtran}

\usepackage{framed,multirow}

\usepackage{amssymb}
\usepackage{latexsym}

\usepackage{url}
\usepackage{xcolor}
\usepackage{amsmath}
\definecolor{newcolor}{rgb}{.8,.349,.1}
\usepackage{booktabs}
\usepackage{lipsum}
\usepackage{wrapfig}
\usepackage{comment}
\usepackage[normalem]{ulem}
\usepackage[inline]{enumitem}
\usepackage{multirow}
\usepackage{balance}
\usepackage{caption} 
\usepackage{subcaption}

\usepackage[linkbordercolor={white},citebordercolor={white}]{hyperref}
\usepackage{graphicx}

\newcommand{\TODO}[1]{{\color{black}#1}} 
\newcommand{\SB}[1]{{\color{black}#1}} 

\newcommand{\pgnt}{PGN++$\mathcal{T}$}
\newcommand{\ours}{\footnotesize{\textit{(Ours)}}}

\newcommand{\eqrefnew}[1]{Eq.~\ref{#1}}
\newcommand{\secref}[1]{Sec.~\ref{#1}}
\newcommand{\tabref}[1]{Tab.~\ref{#1}}
\newcommand{\figref}[1]{Fig.~\ref{#1}}

\begin{document}

\title{PoseGraphNet++: Enriching 3D human pose with orientation estimation}

\author{
 \centering
  \IEEEauthorblockN{Soubarna Banik\IEEEauthorrefmark{1}\IEEEauthorrefmark{2}, Edvard Avagyan\IEEEauthorrefmark{2}, Sayantan Auddy\IEEEauthorrefmark{3} Alejandro Mendoza Garc\'{i}a\IEEEauthorrefmark{4}, Alois Knoll\IEEEauthorrefmark{2}}
  \\

\thanks{\IEEEauthorrefmark{1} Corresponding author: soubarna.banik@tum.de}
\thanks{\IEEEauthorrefmark{2} Technical University of Munich, Germany}
\thanks{\IEEEauthorrefmark{3} University of Innsbruck, Austria}
\thanks{\IEEEauthorrefmark{4} reFit Systems GmbH, Germany}

}

\maketitle

\begin{abstract}
Existing skeleton-based 3D human pose estimation methods only predict joint positions. 
Although the yaw and pitch of bone rotations can be derived from joint positions, the roll around the bone axis remains unresolved.
We present PoseGraphNet++ (PGN++), a novel 2D-to-3D lifting Graph Convolution Network that predicts the complete human pose in 3D including joint positions and bone orientations. 
We employ both node and edge convolutions to utilize the joint and bone features.
Our model is evaluated on multiple datasets using both position and rotation metrics. 
PGN++ performs on par with the state-of-the-art (SoA) on the Human3.6M benchmark.
In generalization experiments, it achieves the best results in position and matches the SoA in orientation, showcasing a more balanced performance than the current SoA.
PGN++ exploits the mutual relationship of joints and bones resulting in significantly \SB{improved} position predictions, as shown by our ablation results.
\end{abstract}


\section{Introduction}
\label{sec:intro}

State-of-the-art (SoA) skeleton-based 3D human pose estimation~(HPE) methods predict only joint positions~\cite{Tekin_BMVC2016_130,mono-3dhp2017,Pavlakos2018,Martinez2017, doosti2020hope,banik_3d_2021,xu2021graph}.
This is not sufficient for understanding the complete body posture, e.g., if the palm is facing up or down, or if the head is facing left or right. 
On the other hand, mesh recovery methods~\cite{kanazawaHMR18, pymaf2021} based on parametric body models~\cite{loper_smpl_2015,SMPL-X:2019} can predict the complete body posture through low-dimensional shape and orientation parameters.
The mesh recovery methods are most suitable in cases such as animation, fashion, etc., where both body shape and bone orientation are required.
However, joint positions and orientations are sufficient to describe the posture in cases where the body shape is not required, such as digital rehabilitation 
and action recognition.
Skeleton-based methods~\cite{Pavlakos2018,Martinez2017, doosti2020hope,banik_3d_2021,xu2021graph} are independent of any body models that are mostly closed-source and require additional large-scale mesh annotations for training~\cite{loper_smpl_2015,SMPL-X:2019}.
This motivates us to enrich skeleton-based 3D HPE by including both joint positions and orientations. 

Skeleton-based 3D pose models benefit from additional structural information~\cite{tome2017lifting}.
To encode the structural information of the human body, recent works~\cite{banik_3d_2021, xu2021graph, zhao_graformer_2021} employ \emph{Graph Convolution Networks}~(GCN).
These methods perform well, but in our view, they have not realized their full potential, as they use only joint (node) features.
The human body can be described as a hierarchical kinematic model, where both the position and the orientation of the parent bone define the child's pose. 
In this paper, we show that in addition to utilizing the structure of skeletal joints, exploiting bone relationships is beneficial for 3D HPE and also helps in learning bone orientations.

We propose \emph{PoseGraphNet++}~(PGN++), a novel GCN that processes 2D features of both joints and bones using our proposed message propagation rules. 
To the best of our knowledge, PGN++ is the first skeleton-based human pose model that predicts both 3D joint positions and orientations.
Our model uses \emph{adaptive adjacency matrices}~\cite{doosti2020hope, banik_3d_2021} and \emph{neighbor-group specific kernels}~\cite{banik_3d_2021,Cai2019} to capture long-range non-uniform relationships among joints, as well as among edges.
PGN++ performs on par with the SoA on the Human3.6M dataset~\cite{ionescu_human36m_2014} in position and orientation metrics.
In generalization performance, it outperforms the SoA on MPI-3DPW~\cite{vonMarcard2018} and MPI-3DHP~\cite{mono-3dhp2017} in terms of position and achieves the SoA performance for orientation on MPI-3DPW~\cite{vonMarcard2018}.
Our results show that it is possible to \emph{lift} 3D orientations from 2D poses through data-driven supervised learning and by enforcing structural constraints.
Our extensive ablation studies show that PGN++ benefits from exploiting the mutual relationships between joints and bones, which leads to improved position predictions.

In summary, our contributions are:
\begin{enumerate*}[label=(\roman*)]
    \item We propose the first skeleton-based pose model that predicts 3D joint positions and orientations simultaneously from 2D pose inputs;
    \item We propose a novel GCN-based pose model that processes both node and edge features using \emph{edge-aware} node and \emph{node-aware} edge convolutions;
    \item We report both position and orientation evaluation on multiple benchmark datasets.
\end{enumerate*}
\section{Related Work}

\emph{Graph Convolution Networks} (GCNs) are deep-learning models for graph-structured data. 
\emph{Spectral graph convolution} assumes a fixed topology~\cite{zhou2020graph} and is suitable for skeleton data.
An approximation of spectral graph convolution, proposed in~\cite{kipfW17}, is widely used in 3D HPE ~\cite{Cai2019, doosti2020hope, banik_3d_2021}. 
Instead of using shared weights as in~\cite{kipfW17}, the benefits of applying different weights for different GCN nodes are shown in~\cite{liu2020comprehensive}.
Our approach is in between~\cite{kipfW17} and \cite{liu2020comprehensive}, and uses different shared weights for different neighbor groups.
Most GCN-based methods utilize either node features~\cite{kipfW17,Cai2019, doosti2020hope,banik_3d_2021} or edge features~\cite{wang2019dynamic}.
The combined use of node and edge features is proposed in~\cite{jiang_co-embedding_2020} for large and sparse graphs.
Our proposed GCN message propagation rules are simplified from~\cite{jiang_co-embedding_2020} to be suitable for small-scale graphs such as the human skeleton.

\emph{3D human pose estimation} methods can be broadly categorized into two groups.
\emph{Direct regression-based} methods~\cite{Tekin_BMVC2016_130,mono-3dhp2017,Pavlakos2018, pavlakos2017coarse} predict the 3D joint locations directly from RGB images.
Methods such as~\cite{Pavlakos2018, pavlakos2017coarse} use a volumetric representation of 3D pose that limits the problem to a discretized space.
On the other hand, \emph{2D-to-3D lifting} methods~\cite{Martinez2017,Pavllo2019,doosti2020hope,banik_3d_2021,xu2021graph,zhan2022ray3d} use 2D poses as input and \emph{lift} them to 3D poses.
Most of these lifting methods~\cite{Martinez2017,doosti2020hope,banik_3d_2021,xu2021graph}, including ours, predict the root-relative pose, whereas the global pose in camera space is predicted in a few~\cite{Pavllo2019,zhan2022ray3d}. 
In terms of architecture, \cite{Martinez2017,zhan2022ray3d} use fully connected networks, and~\cite{Pavllo2019} uses a convolutional architecture. 
PoseGraphNet~\cite{banik_3d_2021}, a GCN-based method, shows the benefit of using the structural constraints of a skeleton.
To capture the long range dependencies between the joints, the adjacency matrix is learnt in~\cite{banik_3d_2021} and attention is used in~\cite{zhao_graformer_2021}.
None of these GCNs~\cite{Zhao2019, doosti2020hope, banik_3d_2021, xu2021graph} extract edge features, and at most use trainable weights for the edges. 
Our approach extracts edge features in addition to node features, which helps to learn the bone orientation and also to improve the position prediction.

\emph{Bone orientation estimation} is also an important part of HPE. 
Two major groups of methods capable of predicting orientation are \emph{Mesh-based} methods and \emph{Non-parametric} methods.
Mesh-based approaches~\cite{li2021hybrik,kolotouros2019spin,yu2021skeleton2mesh}  utilize the low dimensional parameterizations of human body models~\cite{loper_smpl_2015,SMPL-X:2019} and predict bone orientation along with body shape. These methods either derive the mesh parameters directly from RGB images~\cite{kanazawaHMR18,kolotouros2019spin} or use intermediate representations~\cite{li2021hybrik,yu2021skeleton2mesh} for easier optimization.
HybrIK~\cite{li2021hybrik} is a recent method that uses inverse kinematics to infer the mesh parameters while 
SPIN~\cite{kolotouros2019spin} uses an iterative optimization approach.
Skeleton2Mesh~\cite{yu2021skeleton2mesh} is an unsupervised mesh recovery method that retrieves the orientation and body shape from a 2D input pose.
Among the non-parametric models~\cite{luo2018orinet,fisch_orientation_2021}, OriNet~\cite{luo2018orinet} uses 2D keypoint heatmaps and 3D orientation maps for predicting the body pose. 
OKPS~\cite{fisch_orientation_2021} considers some additional virtual keypoints to accurately predict the bone orientations from RGB images.
In contrast to mesh-based methods, which are generally suitable for applications where the body shape is relevant, our focus is only on the 3D pose of the skeleton. Hence, we do not depend on human body models~\cite{loper_smpl_2015,SMPL-X:2019}. 
Our method is closer to the non-parametric methods~\cite{luo2018orinet,fisch_orientation_2021}, but ours is a 2D-to-3D lifting approach, and we enforce the constraints of the human skeleton via graph convolution.
Also, most SoA methods~\cite{kanazawaHMR18,luo2018orinet,kolotouros2019spin,li2021hybrik} that can predict bone orientations only report positional error, whereas our evaluation covers both position and orientation.

\section{Method}

\begin{figure*}[t]
\includegraphics[width=0.95\textwidth]{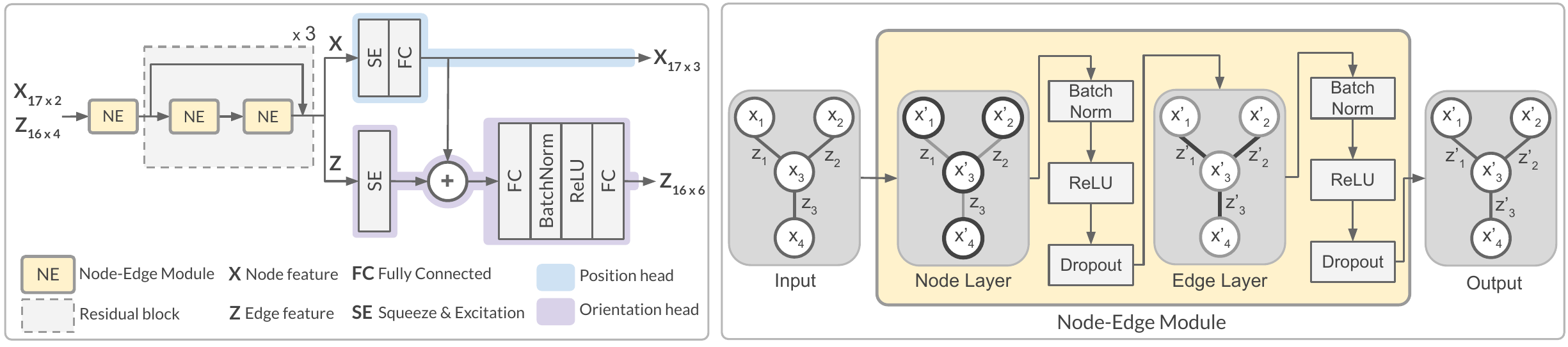}
\centering
\caption{PoseGraphNet++ architecture. (Left) Overall structure of the network. The residual block shown with a dotted box is repeated thrice. (Right) Expanded view of the proposed Node-Edge module, showing the node and edge convolution layers.}
\label{fig:archi}
\end{figure*}

We propose PGN++, a novel GCN for 3D human joint position and orientation regression. %
PGN++ applies the relational inductive bias of the human body through a graph representation of the skeleton.
Unlike other kinematic skeleton-based SoA methods~\cite{banik_3d_2021, xu2021graph, doosti2020hope}, it exploits the mutual geometrical relationship between joint positions and bone orientations.
As shown in \figref{fig:archi},
PGN++ takes the 2D joint coordinates from the image space and the parent-relative 2D bone angles in the form of a rotation matrix as input.
It predicts the 3D position of the joints relative to the root joint (pelvis) in camera coordinate space, and the bone rotations with respect to the parent bone.

\subsection{Preliminaries}
\label{sec:prelim}
In GCN-based 3D HPE~\cite{banik_3d_2021, xu2021graph, zhao_graformer_2021}, the human skeleton is represented as an undirected graph, $G = (V, E)$, where the nodes $V$ denote the set of $N$ body joints, $v_i \in V$, and the edges $E$ denote the connections between them, $(v_i,v_j) \in E$.
The edges are represented by an adjacency matrix, $A \in \mathbb{R}^{N \times N}$.
The graph consists of node-level features, $H \in \mathbb{R}^{N \times F}$, where $F$ is the number of feature channels.
At each layer, the features of the nodes are processed by the propagation rule
$H^{l+1} = \sigma(\tilde{A}H^{l}W^l)$ to form the next layer's features.
Here $\sigma$ is the activation function, $\tilde{A}$ is the symmetric normalized adjacency matrix, $H^l \in \mathbb{R}^{N \times F^l}$ is the node feature matrix of layer $l$, and $W^l \in \mathbb{R}^{F^{l+1} \times F^l}$ is the trainable weight matrix.
$\tilde{A}$ is computed as $\tilde{A} = \hat{D}^{-1/2}\hat{A}\hat{D}^{-1/2},$ where $\hat{A} = A + I$ and $\hat{D}$ is the degree matrix of $\hat{A}$.

\subsection{Proposed Propagation Rules}
\label{sec:prop_rule}
Both joint positions and orientations are important attributes of the hierarchically structured human body.
A parent bone's position and orientation influence the child's position and orientation following forward kinematics.
We exploit these mutual relationships in PGN++ through two types of convolutional layers: a \emph{node} layer and an \emph{edge} layer.
The use of both node and edge convolution in GCN-based pose estimation is novel.
{We simplify the propagation rules of the \emph{edge-aware} node and \emph{node-aware} edge convolutional layers from~\cite{jiang_co-embedding_2020} as:}
\begin{equation}
  \begin{aligned}
H_{v}^{l+1} = \sigma{(\tilde{A}_{v}(T H_e P_e + H_{v}^{l})W_v})\\
H_{e}^{l+1} = \sigma{(\tilde{A}_{e}(T^T H_v P_v + H_{e}^{l}) W_e})
\label{eq:prop_rule}
\end{aligned}
\end{equation}
Here, $H^l$ and $W^l$ have the same meaning as in~\secref{sec:prelim}. The suffixes $v$ and $e$ indicate the node and edge layers respectively.
$N_v$ and $N_e$ are the number of nodes and edges respectively. 
$A_v \in R^{N_v \times N_v}$ is the node adjacency matrix, where each element $A_v(i,j)$ denotes the connectivity of node $i$ and $j$.
Similarly, $A_e \in R^{N_e \times N_e}$ denotes the edge adjacency matrix, where $A_e(i,j)=1$ if $e_i$ and $e_j$ are connected by a node, otherwise 0.
$T \in R^{N_v \times N_e}$ denotes the mapping from nodes to edges, $T_{ve}=1$ if node $v$ is connected to edge $e$.
$P_e$ denotes the weights of the input edge features for the node layer, whereas $P_v$ weighs the neighboring node features for the edge layer.
\SB{Both $P_e$ and $P_v$ are trainable.}
As we operate on non-sparse, small-scale graphs of the human skeleton, our message propagation rules are simplified from~\cite{jiang_co-embedding_2020}, specifically by removing the diagonalization operation and combining the edge and node features prior to the linear combination with weights.
Our node convolutional layer propagates the neighboring nodes' features and additionally the features of the edges connected to the node to form the output node feature.
Similarly, in the edge layer, the features of the neighboring edges and that of the nodes connected to the edge are transformed and aggregated to form the output edge feature of the layer.

\begin{figure}[t]
    \centering
    \includegraphics[width=0.8\linewidth]{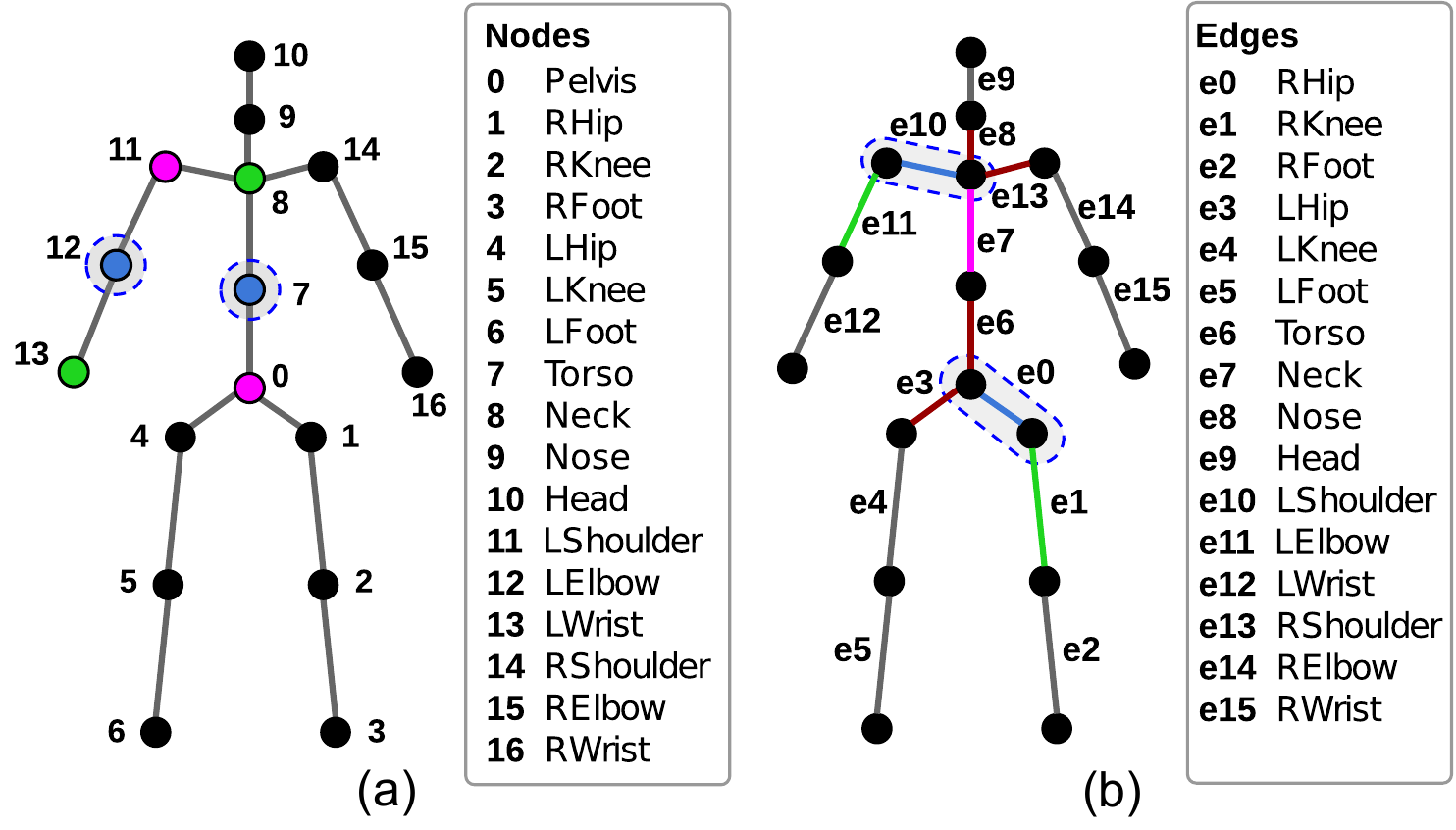}
    \caption{
    Graph representation of the human body: (a) Node definitions with examples of neighbor groups of two nodes (dotted circles). Blue, pink and green show the self, parent and child neighbors respectively. (b) Edge definitions and examples of neighbor groups of two edges (dotted ellipses). Blue, pink, green and maroon highlight the self, parent, child and junction neighbors respectively.
    }
    \label{fig:neighbor_groups}
\end{figure}

\noindent\textbf{Neighbor Groups and Adaptive Adjacency:}
Following PoseGraphNet~\cite{banik_3d_2021}, we apply non-uniform graph convolution by means of neighbor-group specific kernels. It allows extraction of different features for different neighbor groups.
We categorize the node neighbors into three groups:
(i) self, 
(ii) parent (closer to the root joint), and
(iii) child (away from the root joint).
The edge neighbors are divided into four groups: 
(i) self, 
(ii) parent, 
(iii) child, and 
(iv) junction (a special neighbor group, where the graph splits into multiple branches).
\figref{fig:neighbor_groups} shows the graph structure along with some sample nodes/edges and their corresponding neighbors.

{Our final proposed propagation rules (~\eqrefnew{eq:split_kernels}) is extended from~\eqrefnew{eq:prop_rule} to compute the output features of the hidden layers based on the neighbor groups.}
\begin{equation}
  \begin{aligned}
H_{v}^{l+1} = \sigma{(\sum_{g \in G}\tilde{A}^{\dagger}_{vg}(T H_e P_e + H_{v}^{l}) W_{vg}})\\
H_{e}^{l+1} = \sigma{(\sum_{g \in G}\tilde{A}^{\dagger}_{eg}(T^T H_v P_v + H_{e}^{l}) W_{eg}})
\label{eq:split_kernels}
\end{aligned}
\end{equation}
{where $G$ is the set of neighbor groups, $\tilde{A}^{\dagger}_g$ contains the neighbor connections of neighbor group $g$ and $W_g$ is the trainable weight specific to neighbor group $g$.
The $\dagger$ in $\tilde{A}^{\dagger}_g$ indicates adaptive adjacency, which allows learning different contribution rates for different neighbors, unlike $\tilde{A}$ in~\eqrefnew{eq:prop_rule} which enforces equal contribution by each neighbor.}
Adaptive adjacency, as shown in~\cite{banik_3d_2021}, also helps in learning long-range relationships between distant nodes or edges.

\subsection{Rotation Representation}
\label{sec:rot_repr}
Zhou et al.~\cite{zhou_continuity_2019} propose a 6D embedding of rotations, which is continuous in Euclidean space. 
They empirically demonstrate that the use of a continuous representation leads to better performance and stable results.
The 6D representation is achieved by dropping the last column $\mathbf{r}_3$ of an orthogonal rotation matrix $R = [\mathbf{r}_1 \ \mathbf{r}_2 \ \mathbf{r}_3 ]$. i.e. $R_{6D} = [\mathbf{r}_1 \ \mathbf{r}_2]$.
The rotation matrix can be recovered by orthogonalizaton and normalization, specifically by $\mathbf{r'}_1 = \phi{(\mathbf{r}_1})$, $\mathbf{r'}_2 = \phi{(\mathbf{r}_2 - (\mathbf{r'}_1 \cdot \mathbf{r}_2)\mathbf{r'}_1)}$, and $\mathbf{r'}_3 = \mathbf{r'}_1 \times \mathbf{r'}_2$, where $\phi$ denotes the vector normalization operation.
For bone rotations we choose this 6D representation, due to its stable performance during training.

\subsection{Architecture}
PGN++ consists of 3 residual blocks, where each block consists of two \emph{node-edge} (NE) modules
as shown in~\figref{fig:archi}.
An NE module consists of a node and an edge layer, as defined in~\secref{sec:prop_rule}. 
Each of the convolutional layers in NE is followed by a batch normalization layer, a ReLU activation and a dropout. 
A residual connection is added from the block input to the block output.
The residual blocks are preceded by a NE module and succeeded by two downstream modules - the position head and the orientation head.
Both downstream heads contain a \emph{Squeeze and Excitation} (SE) Block~\cite{Hu_2018_CVPR}, followed by one or more fully connected layers.
The SE blocks recalibrate the channel-wise features, $X' = X \odot \mathrm{sigmoid}(W_{ex}\mathrm{ReLU}(W_{sq}X)  ) $, 
where $X$ and $X'$ are the input and output of the SE Block, $W_{ex} \in R^{C \times \frac{C}{d} }$ is the excitation weight, $W_{sq} \in R^{\frac{C}{d} \times C}$ is the squeeze weight, $d$ is the squeeze ratio and \SB{$C$ denotes the input channels}.
The predicted joint positions from the position head are fed to the orientation head.
The concatenated feature is passed to a linear layer, followed by a batch normalization, ReLU and a final linear layer, that produces the bone orientations.

\subsection{Loss Functions}
\label{sec:loss}
The goal of our model is to learn joint locations and bone orientations from the 2D human pose.
Eq.~\ref{eq:loss_mpjpe} shows the Mean Per Joint Position Error (MPJPE) loss function for joint locations that computes the mean squared error between the predicted and the ground truth joint positions. $J^{pred}$ and $J^{gt}$ are the predicted and ground truth 3D joint positions respectively.
\begin{equation}
L_{MPJPE}(J^{pred}, J^{gt}) = \frac{1}{N}\sum_{i=1}^{N} ||J^{pred}_i - J^{gt}_i||_2^2
\label{eq:loss_mpjpe}
\end{equation}
We explore two loss functions to learn the bone orientations: the \emph{Identity deviation (IDev)} loss~\cite{huynh2009metrics} and the \emph{PLoss}~\cite{xiang_posecnn_2018}.
The IDev loss computes the distance of the product of the predicted and the transposed ground truth rotation matrix from the Identity matrix. For two identical rotation matrices, the distance should be zero. \emph{IDev} is defined as:
\begin{equation}
L_{IDev}(R^{pred}, R^{gt}) = \frac{1}{N}\sum_{i=1}^{N} ||I - R^T_{i, gt}R_{i, pred}||_2   
\label{eq:loss_idev}
\end{equation}
where $R^{pred}$ and $R^{gt}$ are the predicted and ground truth rotation matrices.
\emph{PLoss}~\cite{xiang_posecnn_2018} computes the distance between a vector transformed by two rotations and is defined as:
\begin{equation}
L_{PLoss}(R^{pred}, R^{gt}) = \frac{1}{N}\sum_{i=1}^{N} ||R_{i, pred}\hat{X} - R_{i, gt}X||_1
\label{eq:ploss}
\end{equation}
where $\hat{X}$ and $X$ are the predicted and ground truth joint locations.
Empirically, we found that using the L1 norm instead of L2 yielded better results. 
The final loss $L$ is defined as $L = L_{MPJPE} + \lambda L_{Angle}$, where $L_{Angle}$ is either $L_{IDev}$ or $L_{Ploss}$, and 
$\lambda$ is a scalar weight.
For $L_{Angle}$, the predicted 6D rotation is converted to a rotation matrix before computing the loss $L$.
\section{Experiment Setup}

\subsection{Datasets}
\noindent\textbf{Human3.6M} (H36M)~\cite{ionescu_human36m_2014} 
contains 3.6 million images of 7 subjects, from 4 different viewpoints performing 15 indoor actions such as walking, sitting, etc.
The zero rotation pose follows the standard Vicon rest pose~\cite{fisch_orientation_2021} (see \TODO{supplementary materials}).
Following~\cite{Martinez2017,Cai2019,xu2021graph,zhao_graformer_2021}, we use subjects 1, 5, 6, 7, and 8 for training, and subjects 9 and 11 for evaluation.

\noindent\textbf{MPI-3DHP Dataset}~(3DHP)~\cite{mono-3dhp2017} contains 3D poses in indoor and outdoor scenes, but ground truth rotation annotations are not available. We only use this dataset to test the position error.

\noindent\textbf{MPI-3DPW Dataset} (3DPW)~\cite{vonMarcard2018} contains \emph{in-the-wild} 6D poses. 
The joint poses are converted from 3DPW format to H36M format (see \TODO{supplementary materials}).
The zero rotation pose is a T-Pose, unlike H36M.

\subsection{Metrics} 
Following previous work~\cite{banik_3d_2021,xu2021graph,fisch_orientation_2021,yu2021skeleton2mesh}, we compute \emph{Mean Per Joint Position Error} (MPJPE) (protocol 1)~\cite{Martinez2017} for evaluating the joint position prediction.
It is computed as per Eq.~\ref{eq:loss_mpjpe}.
To evaluate the bone orientation prediction, we use two angular metrics - \emph{Mean Per Joint Angular Error} (MPJAE)~\cite{ionescu_human36m_2014} and \emph{Mean Average Accuracy} (MAA)~\cite{fisch_orientation_2021}.
MPJAE is the average geodesic distance $\theta_{sep}$ between the predicted and ground truth bone orientations $R_{pred}$ and $R_{gt}$ respectively, and ranges in $[0, \pi]$ radian. 
MAA~\cite{fisch_orientation_2021} on the other hand is an accuracy metric.
Formally, the metrics are defined as follows
\begin{gather}
    MPJAE = \frac{1}{N}\sum_{i=1}^{N} \theta_{sep}(R_{i, gt}, R_{i, pred})\\
    MAA = \frac{1}{N}\sum_{i=1}^{N} 1 - \frac{\theta_{sep}(R_{i, gt}, R_{i, pred})}{\pi}
\end{gather}
where $\theta_{sep}(R_{i, gt}, R_{i, pred}) = ||\log(R^T_{i, gt}R_{i, pred})||_2$.
The objective is to have lower MPJAE and higher MAA.

\subsection{Implementation Details} 
\label{sec:setup}
\noindent\textbf{Data pre-processing: } We normalize the 2D inputs so that the image width is scaled to $[-1, 1]$. 
We use the camera coordinates for the 3D output and align the root joint (pelvis) to the origin.
The joint angles are provided as Euler angles (\emph{ZXY} format). 
After normalizing the angles to $[-\pi, \pi]$ radian, they are converted to rotation matrices and then to the 6D representation (see~\secref{sec:rot_repr}).
We predict joint positions relative to the root joint and bone rotations relative to the parent.
The global orientation of the root is excluded from the evaluation.
\begin{table*}[t]
  \centering
\caption{MPJPE (mm) of 2D-to-3D lifting methods on H36M under Protocol \#1 (P1). The best and the second best scores are highlighted with bold and underline.}\label{tab:mpjpe}
\resizebox{\textwidth}{!}{\begin{tabular}{lllllllllllllllll}
\toprule
\textbf{Protocol \#1}                                        & Dir.  & Disc. & Eat   & Greet & Phone & Photo & Pose  & Purch. & Sit   & SitD. & Smoke & Wait  & WalkD. & Walk  & WalkT. & Avg. $\downarrow$ \\
\hline
Martinez et al.~\cite{Martinez2017} ICCV‘17 & 51.8  & 56.2  & 58.1  & 59.0  & 69.5  & 78.4  & 55.2  & 58.1   & 74.0  & 94.6  & 62.3  & 59.1  & 65.1   & 49.5  & 52.4   & 62.9  \\
Cai et al.~\cite{Cai2019} ICCV‘19           & {46.5}  & \textbf{{48.8}}  & {47.6}  & {50.9}  & \textbf{{52.9}}  & \underline{61.3}  & 48.3  & \textbf{{45.8}}  & \textbf{{59.2}}  & \underline{64.4}  & \underline{51.2}  & \textbf{{48.4}}  & \underline{53.5}   & \textbf{{39.2}}  & \textbf{{41.2}}   & \textbf{{50.6}}  \\
Zhao et al.~\cite{Zhao2019} CVPR‘19         & 47.3  & 60.7  & 51.4  & 60.5  & 61.1  & \textbf{{49.9}}  & \textbf{{47.3}}  & 68.1   & 86.2  & \textbf{{55.0}}  & 67.8  & 61.0  & \textbf{{42.1}}   & 60.6  & 45.3   & 57.6  \\
Pavllo et al~\cite{Pavllo2019} CVPR‘19      & \underline{47.1}  & 50.6  & 49.0  & 51.8  & 53.6  & 61.4  & 49.4  & 47.4   & \underline{59.3}  & 67.4  & 52.4  & 49.5  & 55.3   & 39.5  & 42.7   & 51.8  \\
Xu et al.~\cite{xu2021graph} CVPR'21 & \textbf{45.2} & \underline{49.9} & \underline{47.5} & 50.9 & 54.9 & 66.1& 48.5 & \underline{46.3} & 59.7& 71.5& 51.4& \underline{48.6}& 53.9& 39.9& 44.1 &51.9 \\
Zhao et al.~\cite{zhao_graformer_2021} CVPR'21 & \textbf{45.2} & 50.8 & 48.0 & \textbf{50.0} & 54.9 & 65.0 & \underline{48.2} & 47.1 & 60.2 & 70.0 & 51.6 & 48.7 & 54.1 & 39.7 & 43.1 & 51.8 \\
Banik et al.~\cite{banik_3d_2021} ICIP'21 & 48.0 & 52.4 & 52.8 & 52.7 & 58.1 & 71.1 & 51.0 & 50.4 & 71.8 & 78.1 & 55.7 & 53.8 & 59.0 & 43.0 & 46.8 & 56.3 \\
\hline
PGN++  \ours{} & 48.9 & 50.1 & \textbf{46.7} & \underline{50.4} & \underline{54.6} & 63.0 & 48.8 & 47.9 & 64.1 & 68.6 & \textbf{50.5} & 48.7 & 53.9 & \underline{39.3} & \underline{42.2} & \underline{51.7}\\
\bottomrule
\end{tabular}
}
\end{table*}

\noindent\textbf{Training details: }We train our model on H36M~\cite{ionescu_human36m_2014} using the Adam optimizer for 20 epochs, using a batch size of 256, and a dropout rate of 0.2.
We use 256 channels in both node and edge layers of PGN++. 
The \emph{MPJPE} and the \emph{IDev} losses are used for training.
The hyperparameters $\lambda$ and the squeeze ratio $d$ are set to 20 and 8 empirically.
The SoA SMPL-based methods~\cite{kanazawaHMR18,kolotouros2019spin,li2021hybrik} predict bone orientation with respect to the T-pose.
In contrast, PGN++ trained on H36M predicts the orientation with respect to Vicon's rest pose.
To be able to compare these methods, we propose a variation of PGN++, named as \emph{\pgnt{}}, which we train on the H36M T-pose annotations provided by~\cite{li2021hybrik}.
Due to fewer samples in this annotation, we train longer (70 epochs) with a batch size 32.

\section{Results}
In this section, we report the position and orientation evaluation of our \SB{baseline} models PGN++ and \pgnt{}. 
Following the SoA methods~\cite{banik_3d_2021,xu2021graph,zhao_graformer_2021}, we use the noisy 2D joint predictions by \emph{Cascaded Pyramid Network} (CPN)~\cite{Chen_2018_CVPR} as the input node features.
For the input bone features, the 2D bone rotations derived from the 2D joint positions are used.

\subsection{Position Evaluation}
We evaluate the predicted positions using the MPJPE (Protocol 1) metric on H36M test set.
\tabref{tab:mpjpe} reports our scores for the 15 actions in H36M along with that of the SoA single-frame 2D-to-3D lifting methods.
PGN++ achieves a mean MPJPE of $51.7$ mm, which is the second-best score overall. 
In inference time, PGN++ outperforms its closest competitors by clocking an average of 3.29 ms for 1000 iterations on an Nvidia RTX 2060 GPU in contrast to 3.36 ms by~\cite{Cai2019} and 6.34 ms by~\cite{zhao_graformer_2021}.
\begin{table}[b]
    \centering
\caption{Position and orientation performance of \pgnt{} and SoA on H36M. Bold and underline have the same meaning as Tab.\ref{tab:mpjpe}.
$*$ indicates orientation evaluation is performed on ten joints following Skeleton2Mesh~\cite{yu2021skeleton2mesh}. 
} \label{tab:oriH36m}
\resizebox{0.95\linewidth}{!}{
\begin{tabular}{l|cc|cc}
\toprule
                                     & \multicolumn{2}{c|}{{Position}} & \multicolumn{2}{c}{{Orientation}} \\
\multicolumn{1}{c|}{{Method}} & {P1} $\downarrow$        & {P2} $\downarrow$      & {MPJAE} $\downarrow$      & {MAA} $\uparrow$      \\
\hline
SPIN~\cite{kolotouros2019spin}                                & 54.88             & \underline{36.65}             & \textbf{0.20}           &       \textbf{93.58}      \\
HybrIK-ResNet (3DPW)~\cite{li2021hybrik}               & \textbf{50.22}             & \textbf{32.38}                 & 0.30           &       90.43       \\
Skeleton2Mesh*~\cite{yu2021skeleton2mesh}                     & 87.1              & 55.4                 & 0.39           &       -       \\
\hline
\pgnt{} \ours{}                      & \underline{50.60}             & 37.21              & \underline{0.23}           &    92.78       \\
\pgnt{}* \ours{}                     & 50.60             & 37.21              & 0.21           &    93.34       \\
\bottomrule
\end{tabular}
}
\end{table}

\subsection{Orientation Evaluation}
\label{sec:ori_perf}
We compare \emph{\pgnt{}} (see sec.~\ref{sec:setup}) with SoA SMPL-based methods~\cite{kolotouros2019spin, li2021hybrik, yu2021skeleton2mesh} that predict both position and orientation. 
As~\cite{kolotouros2019spin, li2021hybrik} do not report their orientation performance, we evaluate these models in our setup.
The MPJAE of Skeleton2Mesh~\cite{yu2021skeleton2mesh} is evaluated on a limited set of 10 joints, and hence we additionally re-evaluate our method on the same joints. 
Among the non-mesh based methods, OKPS~\cite{fisch_orientation_2021} is relevant.
However, we cannot directly compare our method with it, as their code is not available and they report the orientation error only on a subset of joints, details of which are not disclosed in~\cite{fisch_orientation_2021}.
\tabref{tab:oriH36m} reports the position and orientation performance of \emph{\pgnt{}} and the SoA on the H36M test set.
Our method achieves 50.6 mm MPJPE and 0.23 rad MPJAE. 
The yaw, roll and pitch MPJAE are 0.11, 0.13 and 0.13 rad respectively.
\pgnt{} is the second best model in terms of both MPJPE and MPJAE, lagging by only $0.38$ mm and $0.03$ rad behind the best performers HybrIK~\cite{li2021hybrik} and SPIN~\cite{kolotouros2019spin}, respectively.
These results show that the best-performing model in position performs poorly in orientation and vice versa, whereas our method achieves near-SoA performance in both metrics.

\begin{table}[t]
\centering
\caption{Generalization Performance}
\begin{subtable}{0.67\linewidth}
\caption{3DPCK and AUC on 3DHP test set. The best and the second best scores are highlighted with bold and underline.}\label{tab:dhp}
\resizebox{\linewidth}{!}{
\begin{tabular}{ccc}
\toprule
Methods                   & 3DPCK $\uparrow$ & AUC $\uparrow$ \\
\hline
Martinez (H36M)~\cite{Martinez2017}     & 42.5      & 17.0        \\
Yang (H36, MPII)~\cite{Yang20183DHP} & 69.0 & 32.0 \\
Luo (H36M)~\cite{luo2018orinet} & 65.6 &  33.2 \\
Zhou (H36M, MPII)~\cite{Zhou2019HEMletsPL} & 75.3 & 38.0 \\
Zhao (H36M)~\cite{zhao_graformer_2021} & 79.0 & 43.8 \\
Xu (H36M)~\cite{xu2021graph} & 80.1 & \underline{45.8} \\
\hline
PGN++ (H36M)  \ours{} & \textbf{81.1} & \textbf{46.6} \\
\pgnt{} (H36M)  \ours{} & \underline{80.2} & 44.4 \\
\bottomrule
\end{tabular}
}
\end{subtable}
\\
\vspace{3pt}
\begin{subtable}{\linewidth}
\caption{MPJPE (mm) and MPJAE (rad) on 3DPW test set.
Best scores are highlighted in bold.}\label{tab:3dpw}
\resizebox{\linewidth}{!}{
\begin{tabular}{cccc}
\toprule
Methods                   & MPJPE $\downarrow$ & PMPJPE $\downarrow$ & MPJAE $\downarrow$ \\
\hline
HybrIK HRNet~\cite{li2021hybrik} (COCO, 3DHP, H3.6m) & {88.00} & {48.57} & \textbf{0.30} \\
SPIN~\cite{kolotouros2019spin} (H3.6m, 3DHP, LSP, MPII, COCO) & 94.17 & 56.20 & \textbf{0.30} \\
\hline
\pgnt{} (H36M)  \ours{} &  104.96 & 55.30 & \textbf{0.30} \\
\pgnt{} (H36M, 3DHP)  \ours{} &  \textbf{65.64} & \textbf{42.64} & 0.32 \\
\bottomrule
\end{tabular}
}
\end{subtable}
\end{table} 
\subsection{Generalization performance}
\label{sec:generalization}
We evaluate the generalization performance of our models, which are trained on H36M, on the 3DHP~\cite{mono-3dhp2017} and 3DPW~\cite{vonMarcard2018} benchmarks.
Following~\cite{Martinez2017, zhao_graformer_2021}, we report the position metrics 3DPCK and AUC on 3DHP test set in~\tabref{tab:dhp}. 
Both our models outperform the SoA on the 3DHP test set in terms of 3DPCK. 
For AUC, PGN++ performs the best, closely followed by \pgnt{}.

In a separate generalization experiment on 3DPW, we evaluate \pgnt{}, as the orientation annotations of this dataset are with respect to the T-pose.
For a fair comparison with the SoA, which are trained on multiple datasets, we also train \pgnt{} on H36M and 3DHP for this experiment.
We follow a weakly supervised approach, and only use the position loss in case of missing angle labels in 3DHP.
\tabref{tab:3dpw} compares the position and orientation errors of the single- and multi-dataset-trained versions of \pgnt{} with the SoA on 3DPW.
PGN++ (H36M) performs on par with the SMPL-based SoA methods in terms of orientation error, although its position error is high.
3DPW is an outdoor dataset and the viewpoint is very different from that of H36M, which causes the high MPJPE for PGN++ (H36M).
\pgnt{} (H36M, 3DHP) is benefited in this regard from the multi-dataset training and achieves the best MPJPE.
However, since 3DHP does not contain any orientation annotation, the weak supervision of the orientation leads to a drop of 0.02 rad in MPJAE compared to PGN++ (H36M).

\begin{figure}[t]
    \centering
    \includegraphics[width=0.7\linewidth]{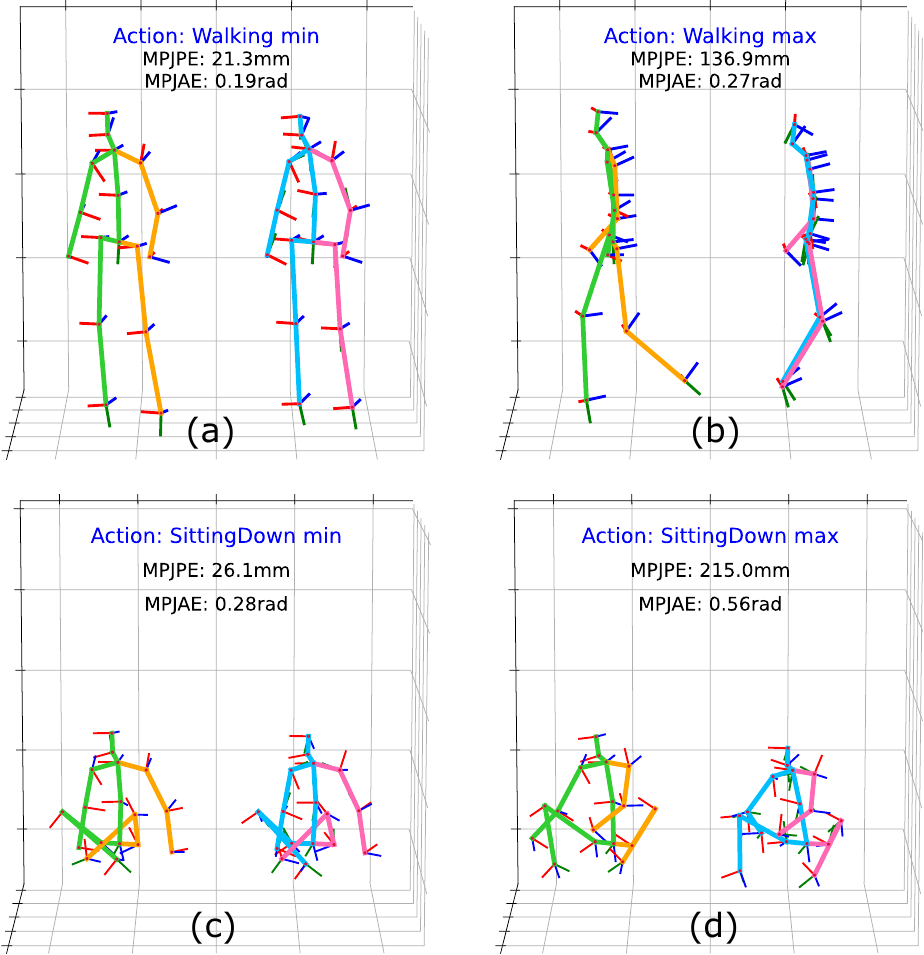}
    \caption{Qualitative results on H36M test set for two
actions with noisy CPN inputs. Each plot shows the ground truth skeleton on the left and the predicted one on the right. (a,c) and (b,d) show the best and failure cases respectively.}
    \label{fig:sample_fig}
    \end{figure}
\subsection{Qualitative Results}
\noindent\figref{fig:sample_fig} shows qualitative examples of minimum and maximum MPJPE for two actions \emph{Walking} and \emph{Sitting down}.
For the best case scenarios (a, c), the overall 6D pose can be recovered with some minor discrepancies (note the head orientation and the roll of the left arm in~\figref{fig:sample_fig}~(c)). 
We observe that the leg pose cannot be recovered for the side view example in~\figref{fig:sample_fig}~(b). 
This arises due to the incorrect 2D prediction made by CPN and does not occur when the ground truth 2D pose is used as input.
Further examples are provided in the \TODO{supplementary materials} and video.

\subsection{Ablation Study}
\label{sec:ablation}


\noindent\textbf{Effect of Node-Edge convolution:}
To test the utility of Node-Edge (NE) convolutions, we compare with a Node-only (NN) version of PGN++.
We change all edge layers to node layers and train using the concatenated node and edge features as input.
The NE model learns the relationships between the joint positions and orientations accurately, and performs better than the NN model on H36M (see~\tabref{tab:NN}). More importantly the NE model generalizes much better on 3DHP.

\noindent\textbf{Multi-task learning:} To analyze the benefits of learning the position and the orientation simultaneously,
we train PGN++ on H36M for the individual tasks using either the positional loss~\eqrefnew{eq:loss_mpjpe} or the orientation loss~\eqrefnew{eq:loss_idev}, and also for both tasks using both losses.
The results are reported in~\tabref{tab:lossfn_study} (sections i, ii). 
When learning both tasks, the errors achieved are lower than that when learning the tasks individually.
This indicates that PGN++ benefits from multi-task learning.

\noindent\textbf{Loss functions:} We evaluate the orientation loss functions IDev~\eqrefnew{eq:loss_idev} and PLoss~\eqrefnew{eq:ploss} in~\tabref{tab:lossfn_study} (sections ii, iii). 
MPJPE + IDev performs the best in both metrics.

\noindent\SB{\textbf{Neighbor-groups and Adjacency Matrices:} We study the effect of \emph{neighbor-group specific kernels} ($W_g$) and \emph{adaptive adjacency matrices} ($\tilde{A}^{\dagger}$) (see~\eqrefnew{eq:split_kernels}). 
~\tabref{tab:adj} reports the results for adaptive and non-adaptive adjacency matrices for neighbor-group specific kernels 
and uniform kernels. 
}
Learning {the node adjacency matrix} $A_V$ improves the positional error significantly (compare (i) with (iii), and (ii) with (iv) in~\tabref{tab:adj}).
Though learning {the edge adjacency matrix} $A_E$ has a marginal effect on the MPJAE, it contributes to reducing the position error (compare (i) with (ii), and (iii) with (iv) in~\tabref{tab:adj}).
{This shows that PGN++ benefits from utilizing the mutual relationships between edges and nodes.}
The neighbor-group specific kernels outperform their counterparts by large margins due to their capability to extract richer neighbor-group specific features.

Based on these ablation experiments, we propose the Node-Edge architecture, MPJPE + IDev loss, 
neighbor-group specific kernels and adaptive adjacency for our PGN++ model.\\


\begin{table}[t]
\centering
\caption{Ablation Studies}
\begin{subtable}{0.7\linewidth}
\centering
\caption{MPJPE and MPJAE of Node-only and Node-Edge \SB{variants} of PGN++(H36M) on H36M and 3DHP test sets.}\label{tab:NN}
\resizebox{\linewidth}{!}{
\begin{tabular}{c|cc|cc}
\toprule
&   \multicolumn{2}{c|}{H36M} & \multicolumn{2}{c}{3DHP} \\
\hline
Model & MPJPE $\downarrow$ & MPJAE $\downarrow$  & PCK $\uparrow$  & AUC $\uparrow$    \\
\hline
Node-only (NN)   & 52.6   & 0.27  &   75.2  & 39.5     \\
Node-Edge (NE) & \textbf{51.7} & \textbf{0.26}  & \textbf{81.1} &\textbf{46.6}   \\
\bottomrule
\end{tabular}
}
\end{subtable}
\\
\vspace{3pt}
\begin{subtable}{0.8\linewidth}
\centering
\caption{MPJPE and MPJAE $\pm$ standard deviation on H36M for different loss functions.}\label{tab:lossfn_study}
\resizebox{\linewidth}{!}{
\begin{tabular}{lccc}
\toprule
& Loss functions              & MPJPE (mm) $\downarrow$ & MPJAE (rad) $\downarrow$ \\
\midrule
\multirow{2}{4em}{(i)} & MPJPE                       & 38.92 $\pm$ 10.69      & 2.20 $\pm$ 0.10  \\
                    & IDev          & 448.10 $\pm$ 26.08     & 0.25 $\pm$ 0.06 \\
\midrule
(ii) & MPJPE + IDev  & \textbf{38.75 $\pm$ 11.13}       & \textbf{0.22 $\pm$ 0.05}        \\
\midrule
(iii) & MPJPE + PLoss               & 39.06 $\pm$ 12.12      & 0.24 $\pm$ 0.04         \\
\bottomrule
\end{tabular}
}
\end{subtable}
\\
\vspace{3pt}
\begin{subtable}{\linewidth}
\centering
\caption{\SB{MPJPE and MPJAE on H36M dataset for adaptive and non-adaptive node ($A_V$) and edge ($A_E$) adjacency matrices, and neighbour-group-specific and uniform kernels. 
}
}\label{tab:adj}
\resizebox{0.9\linewidth}{!}{
\begin{tabular}{ccccccc}
\toprule
&     \multicolumn{2}{c}{Adaptive=(Yes/No)}     & \multicolumn{2}{c}{{Neighbor-group Kernels}} & \multicolumn{2}{c}{{Uniform Kernel}} \\
\hline
Exp. & $A_V$  & $A_E$  & MPJPE $\downarrow$           & MPJAE  $\downarrow$          & MPJPE    $\downarrow$        & MPJAE    $\downarrow$        \\
\hline
(i) & No   & No   & 60.2            & 0.27             & 113.2           &  0.32                \\
(ii) & No   & Yes & 54.0            & 0.27             & 72.7            &  0.31                \\
(iii) & Yes & No   & 52.2            & 0.27             & 82.2            &  0.30                \\
(iv) &Yes & Yes & \textbf{51.7} & \textbf{0.26}      & 79.2            &  0.30          \\
\bottomrule
\end{tabular}
}
\end{subtable}
\end{table}
\section{Discussion and Conclusion}
We present PGN++, the first skeleton-based \emph{2D-to-3D lifting} human pose estimation model that predicts the 3D position and orientation of joints simultaneously.
Our results show that similar to lifting 3D joint positions~\cite{Pavlakos2018,Martinez2017, doosti2020hope,banik_3d_2021,xu2021graph}, it is possible to infer 3D bone orientations from 2D poses by exploiting 2D joint and bone features and the structural bias of the skeleton. 
PGN++ is a light-weight model with 4.6M parameters and has a better inference time than its closest competitors.

SoA mesh-based methods for bone orientation~\cite{kolotouros2019spin,li2021hybrik} only report positional error. 
We evaluate the SoA methods~\cite{kolotouros2019spin,li2021hybrik,yu2021skeleton2mesh} and PGN++ for 
position and orientation error, and
observe that the best performing SoA method for positions 
shows poor performance for orientations, and vice versa. 
In contrast, PGN++ achieves SoA results in H36M considering both metrics.
Our results highlight the importance of a holistic evaluation covering position as well as orientation.
{Importantly, our ablation studies show that learning bone relationships i.e. the edge adjacency matrix, not only facilitates in learning orientation but also improves the position predictions.
This indicates that PGN++ exploits the mutual relationships between nodes and edges and benefits from it.
}
In generalization experiments, PGN++ outperforms the SoA by a large margin in terms of position and is on par with them in terms of orientation.
As our ablation results show, the node-edge convolution is especially helpful in this context over simple node-node convolution.

\SB{While our method has shown promising results, there remains scope for improvement.}
Though infinite 3D poses are possible for a given 2D input, several works~\cite{Pavlakos2018,Martinez2017, doosti2020hope,banik_3d_2021,xu2021graph} including ours show that data-driven supervised 2D-to-3D lifting produces accurate 3D poses. 
Additional information such as joint-specific rotation constraints, camera parameters, or image features can further improve the performance. 
PGN++ also relies on the quality of the input 2D pose. 
To minimize the adverse effects of noisy 2D inputs, we plan to include temporal information in our future work.


\bibliographystyle{IEEEtran}
\bibliography{main}

\end{document}